\documentclass[manuscript]{acmart}

\usepackage[figurename=Figure.]{caption}
\usepackage{cite}
\usepackage{amsmath,amssymb,amsfonts}
\usepackage{graphicx}
\usepackage{textcomp}
\usepackage{xcolor}
\usepackage{algpseudocode}
\usepackage{algorithm}
\usepackage[T1]{fontenc}
\usepackage{amsmath}
\usepackage{esvect}
\copyrightyear{2025}
\acmYear{2025}
\setcopyright{rightsretained}
\acmConference[CSAIDE 2025]{2025 4th International Conference on Cyber Security, Artificial Intelligence and the Digital Economy}{March 07--09, 2025}{Kuala Lumpur, Malaysia}
\acmBooktitle{2025 4th International Conference on Cyber Security, Artificial Intelligence and the Digital Economy (CSAIDE 2025), March 07--09, 2025, Kuala Lumpur, Malaysia}
\acmPrice{}
\acmDOI{10.1145/3729706.3729790}
\acmISBN{979-8-4007-1271-5/25/03}
\begin{document}

\title{Survey of Genetic and Differential Evolutionary Algorithm Approaches to Search Documents Based On Semantic Similarity}

\author{Chandrashekar Muniyappa}
\authornotemark[1]
\email{c.muniyappa@und.edu}
\affiliation{%
  \institution{School of EECS, College of Engineering and Mines, University of North Dakota}
  \city{Grand Forks}
  \country{USA}}

\author{Eunjin Kim}
\email{ejkim@und.edu}
\affiliation{%
  \institution{School of EECS, College of Engineering and Mines, University of North Dakota}
  \city{Grand Forks}
  \country{USA}}
\renewcommand{\shortauthors}{Muniyappa et al.}

\begin{abstract}
Identifying similar documents within extensive volumes of data poses a significant challenge. To tackle this issue, researchers have developed a variety of effective distributed computing techniques. With the advancement of computing power and the rise of big data, deep neural networks and evolutionary computing algorithms such as genetic algorithms and differential evolution algorithms have achieved greater success. This survey will explore the most recent advancements in the search for documents based on their semantic text similarity, focusing on genetic and differential evolutionary computing algorithms.
\end{abstract}


\maketitle

\section{Introduction}
Evolutionary Algorithms (EAs) are a sophisticated class of meta-heuristic methods that emulate the processes of biological evolution. By utilizing mechanisms such as selection, mutation, and crossover, these algorithms effectively tackle a variety of optimization challenges. Different EAs demonstrate distinct capabilities in identifying both local and global optima within expansive solution spaces. Their inherent adaptability allows them to respond efficiently to dynamic changes in their environments, thereby continuously evolving to identify optimal solutions. The increasing adoption of EAs is largely due to their efficacy in addressing complex problems that traditional static methodologies struggle to solve. Notable applications can be observed in fields such as Google Maps and stock market analysis. This research paper will explore the applications of Differential Evolution (DE), Genetic Algorithms (GA), and their variants in measuring text similarity, with the objective of retrieving the most relevant documents based on user input. An overview of the genetic and differential evolution [9] algorithms are shown in ``Algo. 1'' and ``Algo. 2'' respectively.

Text similarity can be defined as the commonness between documents. If there is a greater commonality, then they are similar; dissimilar, otherwise [1]. Similarity can be either lexical or semantic similarity, where lexical similarity refers to character similarity between texts. On the other hand, semantic similarity refers to the actual meaning of words, even if they are different [25]. There are different techniques to represent texts depending on the techniques used to measure the similarity, which is discussed throughout this paper. Although it is the same problem, there are different challenges in measuring the similarity between short and long texts [26]. This survey will focus on how Genetic and Differential evolutionary algorithms can be applied to measure the semantic similarity of documents.

\begin{algorithm}
\caption{Genetic Algorithm (GA) steps}
\begin{algorithmic}[1]
\State \textbf{Initialization:} \textit{Chromosomes representing the initial population can be randomly generated or known values can be used}
\State \textbf{Evaluation:} \textit{Use suitable fitness function to validate the efficiency of the solutions generated}

\State \textbf{Selection:} \textit{The best solutions based on the fitness value will be selected and promoted to next generations. There are various selection strategies like rank, tournament, and roulette wheel selection depending on single or multi objective problem}

\State \textbf{Crossover:} \textit{Crossover or recombination is a process of exchange of information between two parents or chromosomes to produce the best quality offspring. One, Two, and K-Point crossovers and Uniform crossover operators are used in this step.}

\State \textbf{Mutation:} \textit{To promote diversity in selection and avoid getting stuck in the local optima; changes are introduced to chromosomes using mutation operators like bit-flip, Gaussian, Uniform, and so on}

\State \textbf{Replacement:} \textit{Current population will be replaced by new off-springs using Elitism, Steady-state, and General replacement strategies}

\State \textbf{Repeat:} \textit{repeats steps 2 through 5 until satisfactory results are obtained or user specified iterations are reached}
\end{algorithmic}
\end{algorithm}

\begin{algorithm}
\caption{Differential Evolution (DE) algorithm steps}
\begin{algorithmic}[1]
\State \textbf{Initialization:} \textit{Vectors representing the initial population can be randomly generated or known values can be used}

\State \textbf{Mutation:} \textit{Trial vectors are generate using DE/rand/1, DE/best/1, and other mutation strategies with the right 
balance between exploration and exploitation}

\State \textbf{Crossover:} \textit{Crossover or recombination is a process of exchange of information between trial (current) and target (parent) vectors to generate off-spring (new trial) vectors by applying binomial, exponential, and other crossover strategies.}

\State \textbf{Selection:} \textit{The best solutions based on the fitness value will be selected and promoted to next generations.}

\State \textbf{Evaluation:} \textit{Use suitable fitness function to validate the efficiency of the trial vectors generated}

\State \textbf{Repeat:} \textit{repeats steps 2 through 5 until satisfactory results are obtained or user specified iterations are reached}
\end{algorithmic}
\end{algorithm}

\section{Main}
\begin{figure*}
    \centering
    \includegraphics[width=0.8\textwidth]{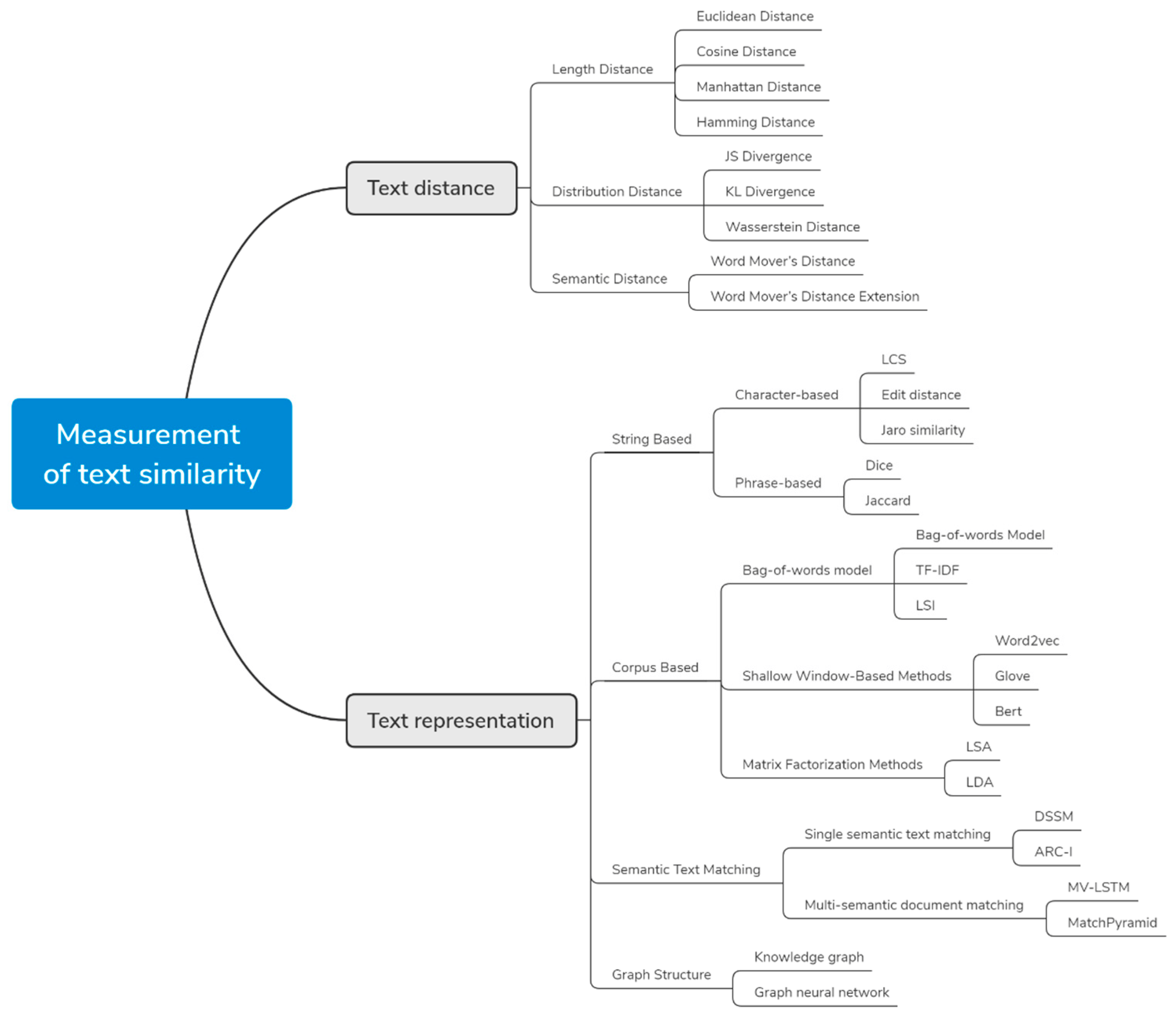}
    \caption{Text similarity measurement (Jiapeng wang et al [1]).}
    \label{fig:similarity}
\end{figure*}

Text similarity plays a crucial role in the efficient identification and categorization of related documents, ultimately enhancing search and ranking functionality. In their comprehensive research, Jiapeng Wang et al [1] studied various techniques for measuring similarity, which are classified based on text distance and representation, as illustrated in Figure \ref{fig:similarity}. Their findings indicate that while string-based methods primarily capture lexical similarity, innovative graph representations provide a more nuanced understanding of contextual meaning. Notably, their research highlights significant opportunities for further exploration of advanced techniques, paving the way for improved document analysis capabilities.   
\par
U. Ahmed et al [3], applied Genetic algorithm to design a agriculture nutrition recommendation system. As part of this research nutrition chemicals were enumerated with integer values and those values were encoded as chromosomes. When the population was initialized randomly they got poor results and the algorithm took more time to converge, hence they initialized the population with nutrient values using the roulette technique. Euclidean distance \ref{euclidean} [13] was used to measure the similarity between the chromosomes as part of fitness functions.

\begin{equation}\label{euclidean}
d_{ij}=\sqrt{\sum_{i=1}^{n} (x_{i}-x_{j}})^2
\end{equation}
Where:
\begin{itemize}
\item ${x_i}$ and ${x_j}$ are two different data points in given sample
\item ${d_{ij}}$ is the distance between them. Lesser this value similar are the two points.
\end{itemize}

Elitism was used for exploitation with one-cross mutation operation. To achieve exploration the best solutions were removed after two generations. Through experiments they prove that results are good.
\par

Bushra Alhijawi et al [4], applied the Genetic algorithm on Movielens movie ratings dataset to build a recommendation system to suggest movies based on similar movie ratings. Vectors of random floating point values were used to capture the similarity between users and movies. The following formula \eqref{bushra} was used to compute users' predictions.

\begin{equation}\label{bushra}
p_x^i=\bar{r_x}+\frac{\sum_{n=1}^{k_x}[sim(x,n)*sim(r_n^i-{r_n})]}{\sum_{n=1}^{k_x}sim(x,n)}
\end{equation}
Where:
\begin{itemize}
\item $\bar{r_x}$ represents the average ratings made by user x for the training items x .
\item $sim(x,n)$ is the similarity value between users x and n. This value is taken from the individual.
\item $r_n^i$ represents the rating of user n on item i.
\item ${k_x}$ is the set of the neighbor to user x. In SimGen, ${k_x}$ is considered to be the set of the training users.
\end{itemize}

The fitness is measured using the Mean Absolute Error between the predicted and known user ratings using the equation \eqref{mae}.

\begin{equation}\label{mae}
MAE=\frac{1}{\#U}\sum_{u=1}^{U}\frac{\sum_{i=1}^{I_U}|p_u^i -r_u^i|}{\#I_u}
\end{equation}
Where:
\begin{itemize}
\item ${\#U}$ represents the number of training users.
\item ${\#I_u}$ represents the number of training items rated by the user u.
\end{itemize}
\par
The research showed improved accuracy and run-time when compared to well know cosine similarity distance measurement. The research makes use of floating point representation which has better performance when compared to the binary representation of chromosomes [5]. However, as described in their paper the floating point values were generated randomly in the range of 0 to 1 and then the Genetic algorithm was used to tune the weights. In this approach, even though two users have the same features, they will be assigned different weights as the weights are generated randomly.  
\par

\begin{figure*}
    \centering
    \includegraphics[width=0.8\textwidth]{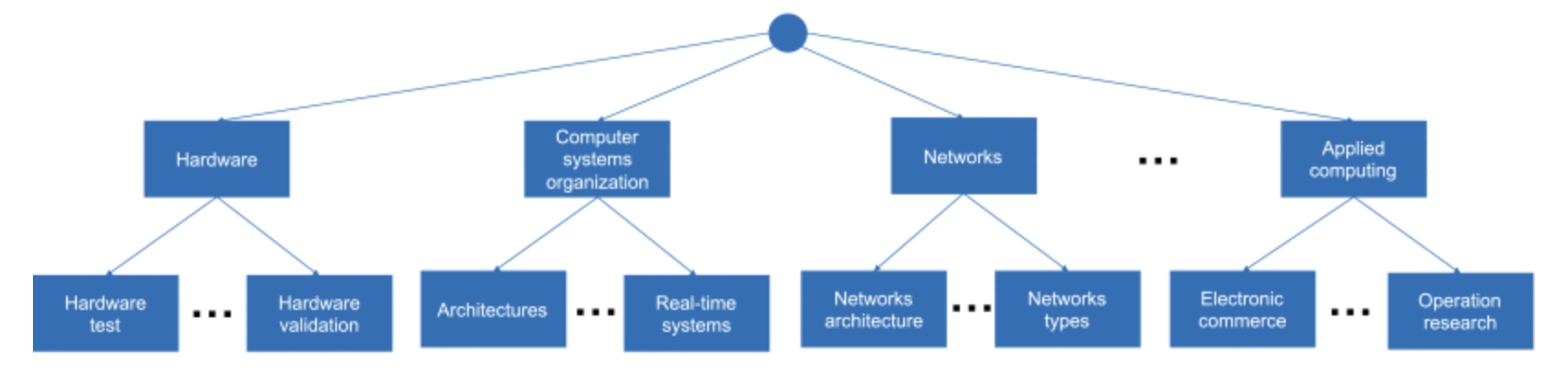}
    \caption{ACM Taxonomy (Alan Diaz-Manriquez et al [6].)}
    \label{fig:img1}
\end{figure*}

Alan Diaz-Manriquez et al [6] conducted a study in which they extracted keywords from research papers and established the ACM taxonomy tree, as illustrated in Figure \ref{fig:img1}. Each category within this taxonomy represents a distinct set of relevant keywords. The taxonomy tree was modeled as a graph using an adjacency matrix, and the minimum distances between nodes were computed utilizing the Floyd-Warshall algorithm [7]. To define the clusters, the researchers enumerated all keywords and selected ${K}$ groups of size ${C}$ as individuals to initialize the Genetic Algorithm (GA). The DB-Index (Davies-Bouldin Index) was used as a fitness function to evaluate the purity of the cluster [15] and guide the optimization process. Furthermore, the algorithm incorporated one-point crossover, random mutation, and tournament parent selection operators to enhance its implementation.

\par

Urszula Boryczka et al. [8] utilized Singular Value Decomposition (SVD), a matrix decomposition technique [10], to model the relationship between users and their movie ratings based on the MovieLens dataset. They designed a movie recommendation system by applying the Differential Evolution (DE) algorithm [9], which they referred to as the "RecRankDE" method. To optimize their solution, they employed the Average Precision (AP) metric, as shown in equation \(\eqref{ap2}\).

\begin{equation}\label{ap1}
P@n=\frac{\textit{Number of relevant items in top n results}}{n}
\end{equation}
Average precision (AP) averages the $P@n$ for different n values:
\begin{equation}\label{ap2}
AP=\frac{\sum_{n=1}^{N}(P@n \times rel(n))}{\textit{Number of relevant items for this query}}
\end{equation}

To evaluate the Top N results fetched, they annotate the results using a binary function ${rel(n)}$ that assigned 1 if the document match is correct and Zero if it was not relevant. The mean of AP was computed across generations for all the users ${U}$ in the dataset.

\par
This is a good approach, they have used an evolutionary algorithm with an efficient floating point representation of chromosomes. Besides, instead of using well-known distance measurement techniques, they are making use of MAP which not only considers the optimal solution but also measures the overall relevance of all the N results. In addition, through experiments, they prove that the results obtained were good. However, the technique described is only for labeled data, the results presented show that the algorithm needs large volumes of labeled data. Also,  they found that when the training set is small the results were poor. 
\par
P. Wang et al [11] combine the niching technique along with Differential Evolution (DE) to select best set of features for a multi-class classification task. Niching is a technique used in multi-objective optimization problem to select the best answer by maintaining balance between exploitation and exploration. In the research, the input features are represented as float point values instead of binary values for better representation; and Hamming distance [16] was used to measure the similarity between the input representations. The standard DE/rand/1 scheme selects the individuals randomly which results in poor performance, to address this issue, they defined a custom mutation operator that takes the distance between the individuals into consideration while selecting the individuals for mutation.  
\par

K. Nandhini et al. [17] implemented a Differential Evolution (DE) algorithm for the purpose of summarizing lengthy texts. The approach began with the preprocessing of sentences, which involved segmenting them into smaller, manageable chunks. Subsequently, the text was tokenized, and Parts of Speech (POS) tagging was applied. Various features of the sentences were analyzed, including position, centrality, similarity, and the percentages of trigger words and hard words. These metrics were incorporated into the DE optimization process to select the most suitable offspring. A Vector Space Model (VSM) was utilized to represent each sentence within a document, with features calculated based on term frequency. Moreover, regression models [18] were employed to generate vectors of weighted feature values for each sentence. The overall sentence score was determined by multiplying the feature vectors by their respective weights, indicating the significance of each sentence within its context. To preserve the original order of the sentences during the mutation process, the sentences were enumerated, and their indexes were used to represent individuals. The information score was derived from the sentence score, as described in \eqref{infoscore}, while cohesion was assessed through cosine similarity [19], as indicated in \eqref{cohesion}. These metrics served as the fitness function to group similar sentences effectively. Custom mutation and crossover algorithms were developed to ensure that the order of sentences remained intact while identifying appropriate points in the chromosomes for these operations.

\begin{equation}\label{infoscore}
F1=\sum_{i=1}^{n}\frac{1}{\sum_{j=i+1}^{n-1}W_i * F_i}
\end{equation}
\begin{equation}\label{cohesion}
F2=\sum_{i=1}^{n}\frac{1}{\sum_{j=i+1}^{n-1} Sim(S_i,S_j)}
\end{equation}
Where
\begin{itemize}
\item ${W_i * F_i}$ is the sentence score calculated using the Feature vector F and linear weights W from the Mathematical regression model for the given ${ith}$ sentence
\item $Sim(S_i,S_j)$ is the pairwise cosine similarity for  ${ith}$ and ${jth}$ sentences.
\end{itemize}

The algorithm will stop when the specified number of iterations are completed. They repeated the steps with the Genetic algorithm to compare the results and performed various statistical analyses on the summary text extracted by both DE and GA implementation and found that the results obtained by DE were better than GA. However, the summary generated was easier to understand when the input text is simple. On the contrary, when the input text is complex and hard to understand due to rare words the accuracy of the summary was reduced. The research used advanced techniques to generate the embedding and effective representation of input data as chromosomes; however, the results deteriorated with advanced English sentences. 

Aytug Onan et al [20], applied the Genetic algorithm to aggregate feature ranks to reduce the number of features in a sentiment classification problem to improve the classification accuracy. When the number of features is large, the model accuracy will be poor due to noise in the features, to overcome this problem, feature importance is ranked and important features are picked. The problem here is, when we rank the feature using different models, multiple features may get the same rank.  If we pick only the top N-ranked features, we will miss all the information from the other features. To solve this problem, they applied the Genetic algorithm to aggregate the top N features from multiple feature selection models to improve the explainability of the final feature set. Finally, validated the effectiveness of the feature set  by using them in a sentiment classification model. To represent chromosomes, a path representation method was used, where each chromosome had a fixed length N, and the position of an element in the chromosome represented the feature rank. However, the default implementations of crossover and mutation operators do not retain the ranking order. Therefore, they used position-based crossover and insertion mutation operators. To measure fitness they used the following formula (8).

\begin{equation}\label{senti}
\delta^* = arg min \sum_{i=1}^{m}W_id(\delta,L_i)
\end{equation}
Where
\begin{itemize}
\item ${\delta}$ is the length of the ${i^{th}}$ list ${L_i}$.
\item ${W_i}$ is the weight representing the importance of features in
 the ${i^{th}}$ list ${L_i}$.
\item ${d}$ is the Spearman foot rule distance function.
\end{itemize}

They applied this approach to multiple open-source datasets and compared it with a regular classification model and found that the results obtained by this approach were far superior. This research clearly demonstrates the drawbacks of the ranking approach, where only top N data points will be picked based on the ranks, and the rest of the elements will be dropped even though they contain meaningful insights. With the help of an evolutionary algorithm, we can capture the latent information from multiple data points through multiple generations without discarding them fully. In addition, it would be interesting to compare the results with Principal Component Analysis (PCA) [23] and advanced neural network auto encoders [24] that are widely used for feature set reduction. On the contrary, even though these models help in improving accuracy, explaining the features that contributed to results would be challenging as most of the features would be combined together. Based on these observations, there is a need to research models that would decode the feature contributions that help in better explainability of the models.
\par

The K-means clustering algorithm [12] is known to have issues with local optima, often resulting in fewer clusters than requested. To tackle this problem, D. Mustafi et al. [21] proposed a hybrid algorithm that combines Genetic Algorithms (GA) and Differential Evolution (DE) techniques. They introduced a novel approach, represented by equation \eqref{centroids}, to select centroids more effectively, rather than relying on random selection. This equation was utilized as the fitness function in their approach.

\begin{equation}\label{centroids}
Max Y=\sum_{i=1}^{K}dis(c_k,C^*)+\sum_{i=1}^{K}\sum_{j=1,i!=j}^{K}dis(c_i,c_j)
\end{equation}
Where
\begin{itemize}
\item ${C^*}$ is the center of the entire solution space.
\item ${c_k}$ is the center of the Kth cluster.
\item ${c_i,c_j}$ are the centers of the ith and jth cluster.
\item ${dis}$ is the cosine similarity measurement.
\item ${K}$ is the required number of clusters.
\end{itemize}

The first part of equation \eqref{centroids} calculates the centroid that is nearest to the majority of the data points in the dataset. The second part computes centroids with diverse features to prevent the model from getting stuck in local optima. The implementation follows a two-phase approach:

${Phase 1:}$ They applied the Genetic algorithm with K-means algorithm using Cosine similarity to measure documents similarities. However, the algorithm starts off by selecting the centroids randomly and then applying equation (9) to build clusters. As the initial step is based on random selection, It will still have all the traditional K-means drawbacks; because a random selection may result in selecting an outlier as a centroid resulting in empty clusters. When this happens, then they apply the Differential evolution K-means algorithm with the following steps as ${Phase 2:}$ to generate the required number of clusters

\begin{itemize}
\item If the requested number of clusters ${k = 2}$ and only one cluster is returned then randomly pick the second one.
\item If the requested number of clusters ${k = 3}$ and only two clusters are returned then the third cluster is returned by vector addition of the first two clusters
${C_3 = C_1+C_2}$
\item If the requested number of clusters ${k > 3}$ and only 3 clusters are returned then the new Centroid is chosen as follows 
${C_j+F \ast (C_k - C_l)}$ where ${C_j, C_k,}$ and ${C_l}$ are the three clusters with the smallest cluster density in the descending order and F is a floating point constant known as mutation factor.
\end{itemize}

These steps are repeated until the desired number of clusters is achieved.The chromosomes are represented using Term Frequency (TF) and Inverse Document Frequency (IDF) encoding of keywords [22]. Experimental results showed that the outcomes produced by the hybrid algorithm are superior to those obtained with the K-means algorithm.
\par

\section{Conclusion}
In this survey, we have studied different techniques used by researchers to apply GA and DE algorithms specifically to measure the semantic similarity of text documents. In addition, we have also shared our observations for each of these studies. The text representation techniques used by researchers can be broadly classified into two types. In the first type, researchers have used basic text representations like Keywords, random number vectors, and graphs with advanced evolutionary algorithms to measure similarity and obtained average performance. In the second type, researchers used advanced techniques to generate text embeddings on their own to apply evolutionary algorithms, and obtained good results. We can observe that there is a relationship between the way texts are represented and the measurement techniques or algorithms applied to capture the semantic similarities between them. Finally, the study found that by applying GA and DE algorithms on regular semantic similarity measurement techniques overall performance can be improved to fetch the Top N records by maintaining right balance between accuracy and relevancy. 

\section{Future Work}
Based on this study we recommend the following future research directions:

\begin{description}
    \item[$\bullet$]None of them used advanced sentence embeddings text representation technique to apply GA and DE algorithms. We have addressed this in one of our research [14] by applying Universal Sentence Encoder embeddings[2]. However, the study was limited to short text or a sentence. More study is required on long texts or paragraphs.
    \item[$\bullet$] Apply GA and DE algorithms on high dimensional solution space, for example: When the text length increases we have to use embedding vectors of larger length like 768, 1024 and so on instead of 512 as in the case of Universal Sentence Encoder to capture the semantic similarity efficiently and evaluate the performance.
    \item[$\bullet$] In the study we found that there is a strong relation between text representation technique and the corresponding similarity measurement technique used. Based on this observation, the traditional distance measurement techniques like Manhattan, Euclidean, Jaccard similarity, to name a few work well with small solution spaces. However, we have to evaluate if they can handle high dimensional vector embeddings and produce correct results with GA and DE algorithms or do we need advanced techniques to capture the interactions between the high dimensional vector embeddings to successful apply GA and DE algorithms.
    
\end{description}

\bibliographystyle{ACM-Reference-Format}

\end{document}